\documentclass[runningheads]{llncs}
\usepackage{times}

\usepackage[T1]{fontenc}
\usepackage{blindtext}
\usepackage{hyperref}
\usepackage{graphicx}
\usepackage{subcaption}

\usepackage{soul}
\usepackage[usenames,dvipsnames]{xcolor}
\definecolor{OliveGreen}{rgb}{0.0, 0.8, 0.0}
\usepackage{xargs}                      
\begin{document}
%
\title{Streamlining the review process: AI-generated annotations in research manuscripts}
\titlerunning{AI-generated annotations in research manuscripts}
%
\author{Oscar Díaz\inst{1}\orcidID{0000-0003-1334-4761} \and
Xabier Garmendia\inst{1}\orcidID{0000-0001-9955-4668} \and
Juanan Pereira\inst{1}\orcidID{0000-0002-7935-3612}}

\authorrunning{Díaz, Garmendia and Pereira}

\institute{University of the Basque Country (UPV/EHU), San Sebastián, Spain\\
\email{\{oscar.diaz,xabier.garmendiad,juanan.pereira\}@ehu.eus}}

%
%
%
%
\maketitle              
\begin{abstract}
The increasing volume of research paper submissions poses a significant challenge to the traditional academic peer-review system, leading to an overwhelming workload for reviewers. This study explores the potential of integrating Large Language Models (LLMs) into the peer-review process to enhance efficiency without compromising effectiveness. We focus on manuscript annotations, particularly excerpt highlights, as a potential area for AI-human collaboration. While LLMs excel in certain tasks like aspect coverage and informativeness, they often lack high-level analysis and critical thinking, making them unsuitable for replacing human reviewers entirely. Our approach involves using LLMs to assist with specific aspects of the review process. This paper introduces \textit{AnnotateGPT}, a platform that utilizes GPT-4 for manuscript review, aiming to improve reviewers' comprehension and focus. We evaluate \textit{AnnotateGPT} using a Technology Acceptance Model (TAM) questionnaire with nine participants and generalize the findings. Our work highlights annotation as a viable middle ground for AI-human collaboration in academic review, offering insights into integrating LLMs into the review process and tuning traditional annotation tools for LLM incorporation.

\end{abstract}

\section{Introduction} The growing number of research papers being submitted to academic journals presents a substantial challenge to conventional peer-review system \cite{PublonsG57}. This trend contributes to a significant rise in the workload for peer review: over 15 million hours are dedicated to reviewing manuscripts \cite{PeerRevi42}, while academics are handling an average of 14 manuscript reviews each year, with each review requiring about 5 hours \cite{Ware2011}. This burden frequently leads to reviewing being undertaken `without sufficient care' \cite{spyns2015scientific}. This calls for assistance in reconciling efficiency and effectiveness in peer review.

Recent advancements in generative AI, especially Large Language Models (LLMs), hold some promise for alleviating the review load \cite{srivastava2023day,checco2021ai,lin2023automated}. However, there are notable issues with the standalone AI-generated reports \cite{yuan2022can}. Firstly, these reports often lack high-level analysis, failing to provide a comprehensive and insightful evaluation. Secondly, the style adopted by the generated reports tends to imitate rather than demonstrate critical thinking and independent judgment. Finally, a prominent shortcoming is the absence of questioning and probing, which are crucial for fostering rigorous scholarly discourse.  However, Yuan et al. acknowledge that in certain areas, such as aspect coverage and informativeness, AI systems can surpass human reviewers \cite{yuan2022can}. This moves the role of LLMs to complement human reviewers rather than replace them. This calls for a nuanced approach in which LLMs assist reviewers with specific and tangible aspects of the review process, rather than expecting LLMs to handle the entire report. This work focuses on manuscript annotations, particularly excerpt highlights, as the middle ground for reliable AI-human collaboration.

Highlighting is a longstanding practice that helps readers comprehend and understand the material. When relevant information is already highlighted, it provides visual cues that guide the reader's attention and improve comprehension \cite{donnel2004}. LLMs are considered proficient in identifying relevant information, i.e., annotation \cite{yuan2022can}. By outsourcing annotation to LLMs, reviewers can benefit from departing from an annotated manuscript. This not only provides a head start, but the expectation is to attain the same benefits as for human-annotated manuscripts, i.e., to guide the reviewer's attention and improve comprehension, ensuring that no valuable excerpts of the manuscript are overlooked. This leads to our research questions:
\begin{quote}
   RQ1: \textit{How could the task of LLM-authored annotations be integrated into the review process?} \\
   RQ2: \textit{How could traditional annotation tools be tuned for LLM-assisted academic review?}
\end{quote}

In answering these questions, we contribute by
\begin{itemize}
    \item making a case for annotation as a suitable middle ground between human reviewers and LLM interaction. (Section \ref{sec:practice}).
    \item providing proof-of-concept through \textit{AnnotateGPT}, a dedicated platform for manuscript review with GPT-4 as the LLM (Section \ref{sec:intervention}).
    \item evaluating \textit{AnnotateGPT}  through a TAM questionnaire (n=9) and generalizing the results to other stakeholders (Section \ref{sec:tam} \& Section \ref{sec:generalization}).
\end{itemize}

\section{Related work}\label{sec:relatedWork}
AI interventions in peer reviewing can be framed along two dimensions. The first dimension is the targeted stakeholder: the editor, the reviewer, or the author. The second dimension is the role played by the AI: AI for automation (i.e., AI systems that replace human work) and  AI for augmentation (i.e., AI systems that integrate with human expertise to improve decisions)  \cite{enholm2022artificial}.  This section is structured along with these dimensions.

\paragraph{The editor as the stakeholder}
Goshal et al. target editors by proposing a sentiment analyzer to automatically validate a research manuscript based on both the manuscript and the reviews from reviewers to predict the acceptance decision. They claim their system outperforms existing baselines, offering an enhanced level of confidence for editors, particularly in cases where reviewers are unresponsive \cite{ghosal2019deepsentipeer}. In contrast, Checco et al. introduce a regular neural network model trained with 3300 manuscripts, which demonstrates the ability to accurately predict the outcome of the peer review process based on superficial features of the manuscript. The authors acknowledge the importance of addressing potential biases and ethical concerns associated with these tools \cite{checco2021ai}.

\paragraph{The reviewer as the stakeholder}
Wang et al. introduce \textit{ReviewRobot}, an automated tool for assigning review scores and generating comments on aspects like soundness and novelty. Utilizing Gated Recurrent Unit (GRU) models, the tool generates these evaluations based on knowledge graphs created from manuscripts using information retrieval techniques. In tests within NLP and ML domains, \textit{ReviewRobot} demonstrated an accuracy range of 71.4\% to 100\% \cite{wang2020reviewrobot}. Alternatively,  Yuan et al., introduce \textit{ReviewAdvisor}. This tool leverages a model trained in abstractive summarization to produce reports, evaluated across seven distinct criteria \cite{yuan2022can}. As to resorting to LLMs, Biswas et al. investigated ChatGPT's effectiveness as a standalone AI reviewer for academic journals. They fine-tuned ChatGPT using journal-specific guidelines, then had it review a case report, suggesting revisions. The AI's performance, notably in identifying methodological flaws and providing insightful feedback, was closely aligned with human reviewers. This method, especially beneficial for preprint servers, highlights ChatGPT's potential in improving unreviewed manuscript quality \cite{biswas2023focus}. Conversely, Srivastava et al. introduce a two-step procedure for using ChatGPT to \textit{automatically} review manuscripts. First, the manuscript’s content is inputted into ChatGPT, with a request to provide a summary of the manuscript and assess various aspects like clarity and novelty. Then, the output is passed to ChatGPT for sentiment analysis, which contributes to the decision-making of paper acceptance or rejection \cite{srivastava2023day}.

In contrast to \cite{ghosal2019deepsentipeer,checco2021ai,wang2020reviewrobot,yuan2022can}, the key difference in our work lies in how the technology is applied. Our LLM-based approach does not seek to replace human judgment but to utilize the advanced capabilities of LLMs to offer more profound, context-sensitive insights.

Most of the previous works are arranged along the `AI for automation' quadrant: the human is not in the loop but at the end of the loop.  By contrast, we focus on `AI for augmentation', i.e., we do not aim at replacing human judgment but to utilize the advanced capabilities of LLMs to offer more profound, context-sensitive insights. The ultimate rationale for an augmentative rather than an automation approach to manuscript review is ethical. Without human supervision, LLMs raise important ethical issues \cite{park2023use}. In line with these considerations, Biswas et al. conclude that ``a balanced approach that combines the strengths of AI with human expertise is key to achieving the best outcomes in the peer-review process'' \cite{biswas2023focus}. 

The question arises as to how to find this `balanced approach'. We advocate for annotations to serve as this middle ground. Here, annotation is seen as a preliminary step before the review process. Instead of starting from scratch, the reviewer will work with a manuscript that already has annotations made by the LLM. This approach offers several time-saving benefits, as important excerpts can be located easily, while still allowing the reviewer to maintain control over the interpretation and value of the highlighted sections. Annotations should not be viewed as a justification for not reading the entire paper, but rather as a means of highlighting paragraphs that may be overlooked when reading under stressful conditions. This approach is less reliant on technology and helps to address ethical concerns, as the LLM's role is limited to identifying important sections. In order to incorporate LLMs into the reviewing process, it is necessary to first define this process. This brings us to the next section.

\section{RQ1: How could the task of LLM-authored annotations be integrated into the review process?} \label{sec:practice}

Peer review is not solely focused on determining whether a manuscript should be accepted or rejected. It also serves the purpose of enhancing the quality of the published paper by offering valuable feedback. Approximately 90\% of researchers believe that the primary benefit of peer review lies in its capacity to provide constructive criticism, thereby improving the final version of the paper \cite{ware2011peer,yuan2022can}.  
This places peer-review as a feedback practice.  

Feedback provision is being extensively investigated in education. 
Feedback on student assignments is pedagogical, and aimed at enhancing learning, understanding, and skill development. Alternatively, peer review in research manuscripts is evaluative, intended to uphold and improve the quality and integrity of scholarly work before publication. That said,  both practices aim to provide constructive criticism to help the author(s) improve their work, whether it be a student’s essay or a researcher’s manuscript. Good practice in both involves providing specific and actionable advice, pointing out not just what is lacking but also how to improve it. 
Nicol provides ten recommendations for constructive feedback for textual assignments \cite{Nicol2010}.  We focus on three of these recommendations related to the substance of feedback rather than to its linguistic aspects (e.g., understandability or politeness). These include ensuring that feedback is contextualized, specific, and timely. The rest of this section delves into each of these aspects. Bold font is used for cues for later functional requirement.

\textit{Contextualized feedback} refers to feedback being framed about the learning outcomes \cite{Nicol2010}. This implies the existence of a review framework that guides the reviewer through the reviewing process. However, reviewing is a diverse practice \cite{tennant2017multi}, with communities and research areas requiring different aspects to be assessed. While student feedback is often more directive, with specific suggestions for improvement and learning, peer review is more critical, with an emphasis on identifying gaps, questioning methodology, and ensuring the validity of the findings. This calls for \textbf{making explicit the review criteria}. Additionally, contextualized feedback might not be limited to the review criteria but also include \textbf{the sentiment of the feedback}, i.e., positive remarks versus negative remarks,  on the pursuit of balanced feedback that helps to motivate rather than discourage.

\textit{Specific feedback} refers to avoiding general comments and instead \textbf{being more accurate about the excerpts in the manuscript where the feedback applies} \cite{Nicol2010}. This is commonly achieved by annotating the manuscript while reading, and then, quoting when writing the review. Similar to ``grammatical typos'', specific feedback should include the excerpts where the ``argumental typos'' or ``methodological typos'' apply.

\textit{Timely feedback} highlights the fact that feedback given close to the time of action reinforces either the correctness of the action or the need for change \cite{Boud2013}. This issue might be better captured by the term ``performant feedback'' as a cause previous to Nicol's timely feedback. In software development, a ``performant'' application (reviewer) would respond to user inputs quickly (journal editor, PC chair), processes data efficiently (manuscript review), does not consume excessive computational resources (reviewer's time), and can handle high volumes of work without degradation in performance (reviewer's focus). This raises the question of how to enhance the performance of reviewers. This is when LLM annotation comes into play.

Outsourcing the annotation to LLMs could help meet the quality criteria mentioned above. Specificity could be achieved by automatically integrating the excerpts into the review. Contextualization could be achieved by color-coding the excerpts, with each color representing a different criteria for the review. Lastly, timeliness is related to the improvement in focus that is expected to be achieved when reading annotated manuscripts (w.r.t. un-annotated manuscripts)  \cite{donnel2004}.

\begin{figure}[!h]
    \centering
    \includegraphics[width=1.05\textwidth]{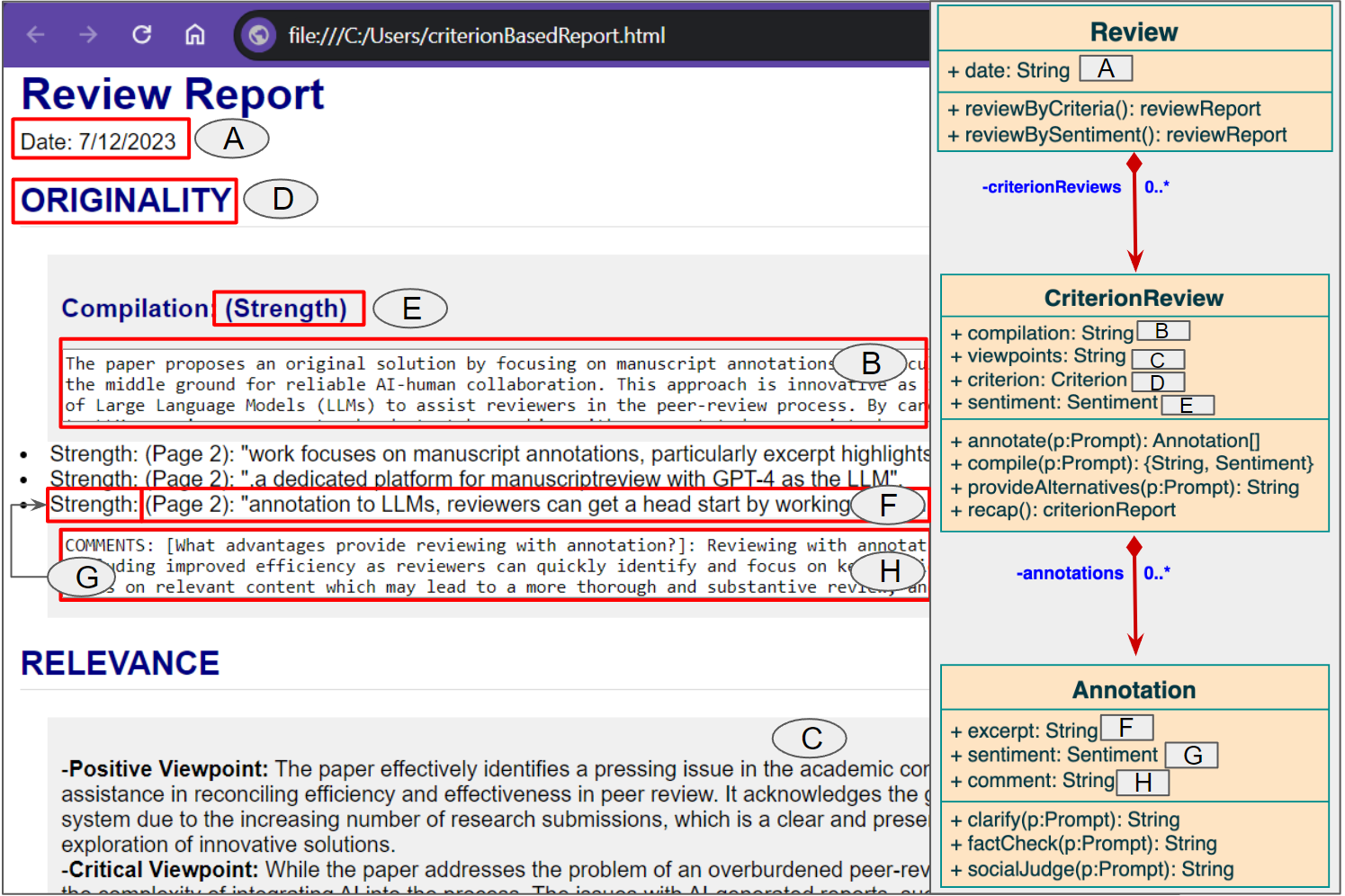}
    \caption{A review report and its UML conceptualization}
    \label{fig:uml}
\end{figure}

LLMs can be useful not only for annotating but also for assembling the final review.
A review is more than just a collection of annotations. Annotations provide the basis for evaluating how well the manuscript meets the review criteria. Criterion-driven revision is conceived as a gradual and iterative process wherein frequent back-and-forth through the manuscript is expected. The aim is to assess the extent to which the authors' claims meet criteria such as originality, relevance, rigor, etc. Fig. \ref{fig:uml} illustrates this vision and serves as a requirement document for dedicated platforms for peer review. The implications are twofold.

Structurally, a \textit{Review} is composed of multiple  \textit{CriterionReviews}. Each \textit{CriterionReview} is based on a set of \textit{Annotations}. \textit{Review, CriterionReview}, and \textit{Annotation} all have attributes that keep their state. In the case of annotations, the state is the \textit{excerpt} that is being highlighted, which can be accompanied by \textit{comments} and assigned a \textit{sentiment} (such as strength or weakness). For \textit{CriterionReview}, the attributes \textit{compilation} and \textit{viewpoints} keep strings that provide different ways of summarizing a review using its annotations. Lastly, a review includes a review report, which can be structured based on either the revision criteria (such as rigor or originality) or the sentiment (strength or weakness).

Behaviorally, the \textit{Review} is created gradually from the bottom-up, beginning with annotations and criterion-based reviews. Eventually, it is compiled into a structured review based on either criteria or sentiment. In other words, the revision process is conducted gradually. The reviewer leads the process and may utilize the LLM to identify relevant excerpts related to a specific review criterion. Subsequently, the reviewer can elaborate on these highlights by requesting additional comments or providing their own.
Annotations and comments are eventually \textit{compiled}, and the LLM can be used to provide different \textit{viewpoints}. Fig. \ref{fig:uml} makes explicit the participation of the LLM by noting as \textit{Prompt} the implementation for those functions. 

Returning to RQ1, the dialogue between the LLM and the reviewer is organized based on the reviewing criteria. The reviewer has the ability to request annotations for a particular criterion, which can later be fact-checked or social judged. Additionally, the reviewer can supplement the LLM annotations with their own comments and sentiments. Finally, the LLM can assist in compiling the final review by incorporating the annotations (no matter their origin). By including these annotations, aligning them with the reviewing criteria, and automatically generating the review, we aim to provide specific, contextual, and timely feedback in the review process. The question remains about how this process can be accommodated within traditional annotation tools.

\section{RQ2: How could traditional annotation tools be tuned for LLM-assisted academic review?}\label{sec:intervention}

A review platform is a dedicated interface for manuscript reviewing. We consider the platform to be ``AI-powered'' if it incorporates an LLM into the process. This pertains to the way in which the LLM prompts are composed, invoked, sequenced, and integrated into the platform's interface. For example, ChatGPT primarily engages users through a question-answer format  \cite{OpenAI2023ChatGPT}. Alternatively, we take a different approach by using annotations as the main form of interaction between reviewers and the LLM. This shift from a linear dialogue to an annotation-based approach allows for a more layered and nuanced interaction with the manuscript. Therefore, the main contribution of this work lies not so much in the prompting (which follows traditional practices, as discussed later), but in the proposed interaction model for LLM-human collaboration\footnote{ However, we should not overlook the importance of predefined prompt templates. These templates not only make it easier for non-technical users to interact with the model by lowering technical barriers, but they also improve interaction efficiency. Users can simply select appropriate prompts from the existing templates instead of having to create their own from scratch. Moreover, since these templates are carefully designed and tested, they tend to be more precise and reliable, minimizing errors that may result from unclear or vague user instructions.}. This section presents a proof-of-concept:  \textit{AnnotateGPT}. 

  \textit{AnnotateGPT} is a browser extension on top of the PDF visor for Chrome. 
  \textit{AnnotateGPT}  applies a dynamic overlay to the PDF manuscript, where creation, updating, rendering, and assembly of annotations to conform the final review is conducted with the discretionary participation of GPT-4.  \textit{AnnotateGPT}  is available for public download at the Chrome Web Store\footnote{\href{https://rebrand.ly/annotateGPT}{https://rebrand.ly/annotateGPT}. APIkey for GPT-4 is required}, and the source code is publicly available under the MIT license on GitHub \footnote{\href{https://github.com/onekin/AnnotateGPT}{https://github.com/onekin/AnnotateGPT}}. A video  is available on YouTube\footnote{\href{https://rebrand.ly/annotateGPT\_Video}{https://rebrand.ly/annotateGPT\_Video}}. As additional evidence, excerpts from this very manuscript have been highlighted using \textit{AnnotateGPT} along two review criteria: problem relevance and originality with a yellow and green highlighting, respectively. 


\textit{AnnotateGPT}  makes the manuscript available to GPT-4. 
It is most important to note that this manuscript is only held temporarily to provide the necessary service, such as generating the annotations. After the session ends or when the data is no longer needed for the immediate task, the files are deleted and not retained.  Next, we delve in how \textit{AnnotateGPT} accounts for the different entities in Fig. \ref{fig:uml}. \textit{AnnotateGPT}'s default prompts can be found here*.

\subsection{ Annotation }
\textbf{Create}. Annotations are created by prompting GPT-4 through the \textit{annotate} prompt. We utilize the "Reverse Prompt Engineering" technique as our core design strategy  \cite{HowtoMas25:online}. This technique involves presenting the language model with desired outputs, along with examples, and instructing it to generate corresponding prompts iteratively. The output is JSON formatted, as described later on. The ``\textit{annotate}'' prompt has a single parameter: the review criterion. Besides the criterion, the prompt includes a request for (1) three excerpts that provide the evidence that sustains the criterion in the paper and (2) the analysis of the sentiment to determine whether each excerpt meets the criterion.\footnote{The number of excerpts that can be returned is configurable. Based on our experience, we have found that the evidence tends to become faint beyond the second excerpt.}.

\begin{figure}[!h]
    \centering
    \includegraphics[width=1.0\textwidth]{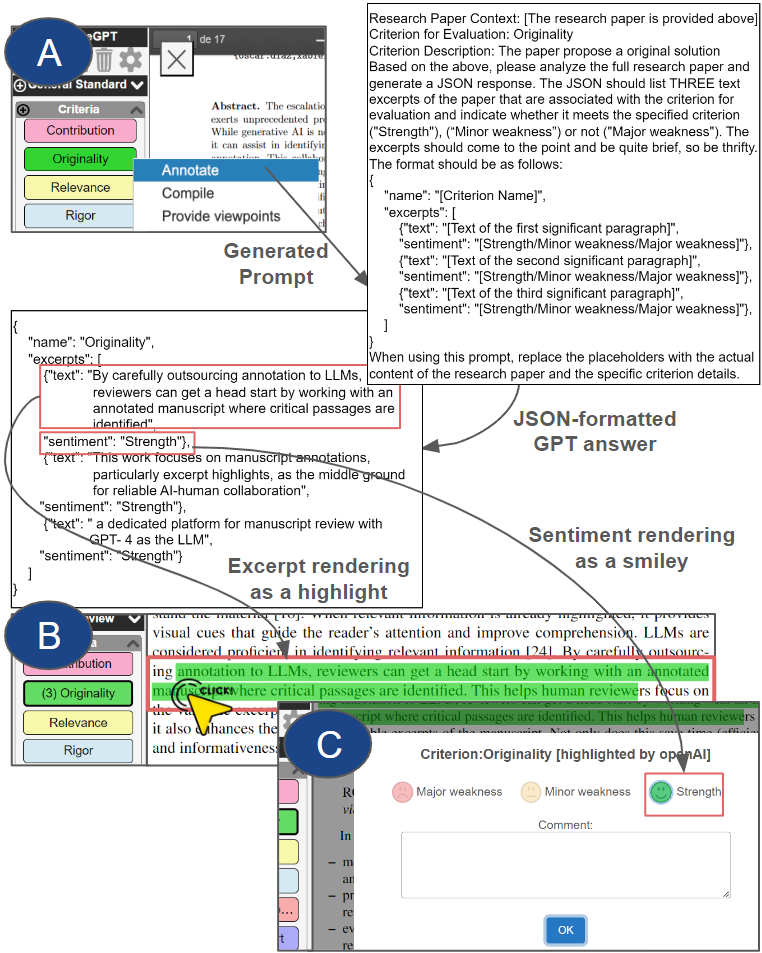}
    \caption{Creation of annotations }
    \label{fig:promptRendering}
\end{figure}

 The \textit{annotate} function belongs to the \textit{criterionReview} class. This class has a highlighter bar as its GUI counterpart (see later). Fig. \ref{fig:promptRendering} outlines the interaction sequence. First, the reviewer should click on the highlighter criterion to display a contextual menu. By selecting \textit{Annotate} (A), a prompt will be generated and sent to GPT-4. GPT-4 should be instructed to format the answer in JSON. The returned JSON should then be rendered, with excerpts located and color-coded highlighting (B), and sentiments reflected through smileys (C).

\begin{figure}[htbp]
\centering
\begin{subfigure}[b]{\textwidth}
    \includegraphics[width=\textwidth]{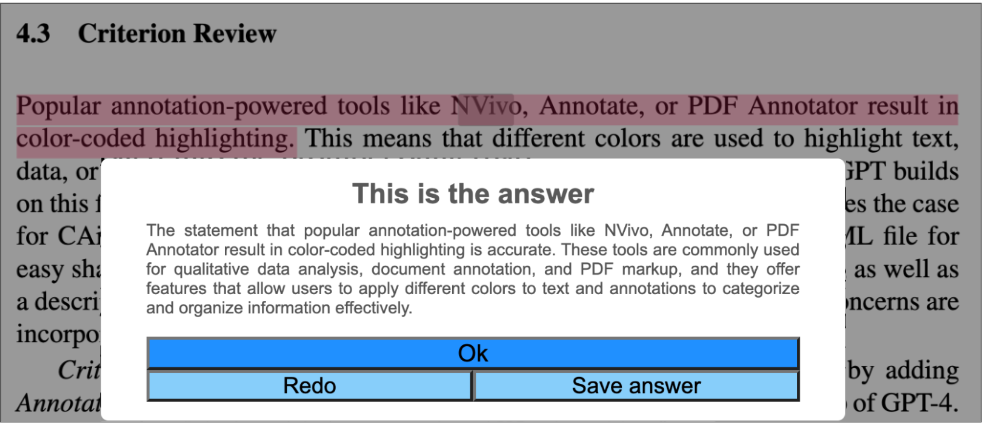}
    \caption{Fact Checking on color-coding support}
    \label{fig:factChecking}
\end{subfigure}

\begin{subfigure}[b]{\textwidth}
    \includegraphics[width=\textwidth]{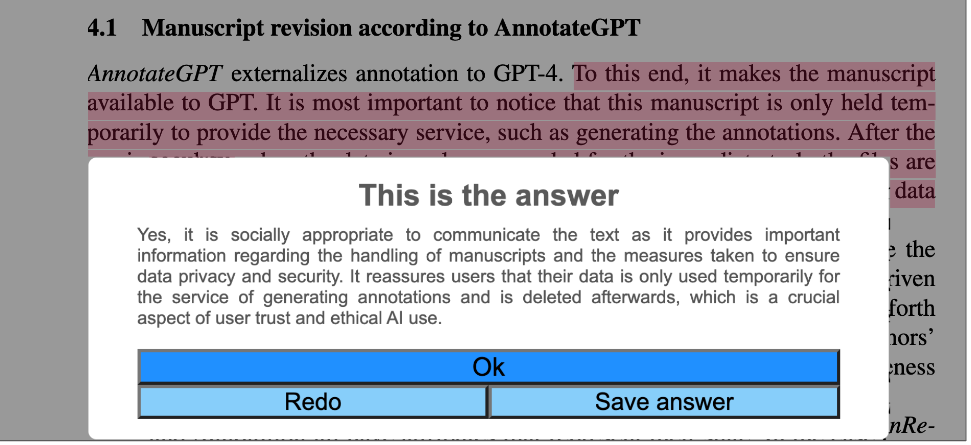}
    \caption{Social Judgement on confidentiality concerns}
    \label{fig:socialJudgement}
\end{subfigure}

\begin{subfigure}[b]{\textwidth}
    \includegraphics[width=\textwidth]{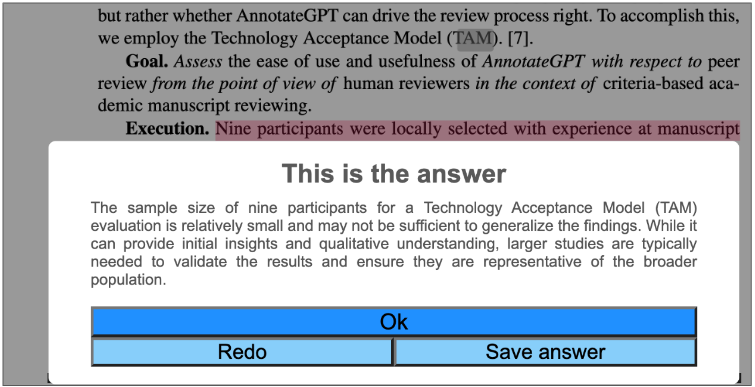}
    \caption{Clarification request through the prompt: `\textit{9 subjects enough for TAM evaluation?'}}
    \label{fig:clarify}
\end{subfigure}
\caption{Annotation-centric prompting  using annotations from  this very manuscript  }
\label{fig:combinedFigures}
\end{figure}

\textbf{Update}. If annotations are created through prompting, it seems reasonable to assume that updates might also be prompt-driven. Specifically,  AnnotateGPT accounts for requesting additional evidence to fact-check the text in the annotation, address social concerns, or ask open-ended questions to clarify doubts about the annotation content. These operations are available through the contextual menu when on an annotation. Fig. \ref{fig:combinedFigures} illustrates these three scenarios for this very manuscript. If appropriate, the output can be saved for later compilation. 

While AnnotateGPT is not intended for the reviewer to conduct the annotation themselves, but rather as a result of prompting, the reviewer is also able to create their own annotations by selecting excerpts from the manuscript and clicking on the associated criterion. This can be useful to highlight excerpts that the LLM failed to identify or that are deemed important for later revision.

\subsection{Criterion Review}
Annotation-powered tools like NVivo, Annotate, or PDF Annotator result in color-coded highlighting. This means that different colors are used to highlight text, data, or other information based on specific criteria or categories. AnnotateGPT builds on this familiarity, where colors now represent review criteria.
 Fig. \ref{fig:promptRendering}  illustrates the case for Contribution, Originality, Relevance and Rigor. 
Review criteria can be introduced manually or through an XML file for easy sharing. Each criterion includes a name (to be displayed in the sidebar), as well as a description and, optionally, actionable recommendations to assess the criteria. These concerns are incorporated into the prompt counterpart.

\begin{figure}[!h]
    \centering
    \includegraphics[width=1\textwidth]{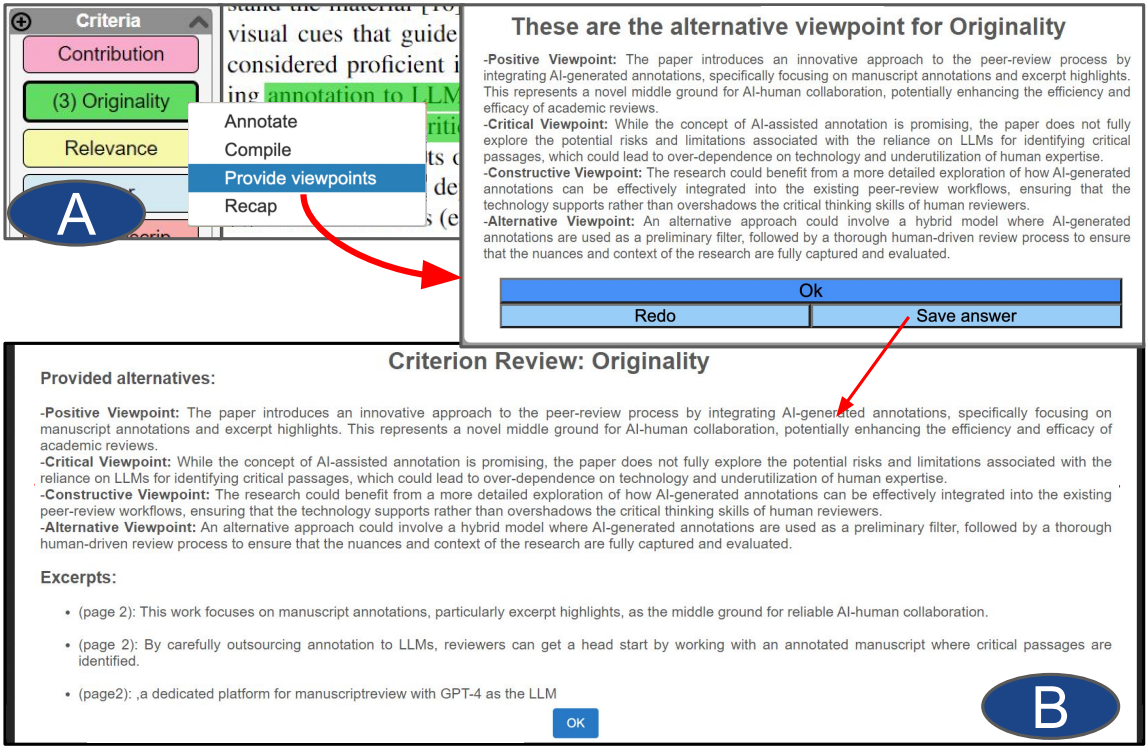}
    \caption{Viewpoints for CriterionReview: the prompt includes annotations collected for the criterion at hand}
    \label{fig:viewpoint}
\end{figure}

\textit{CriterionReview} instances are initially empty, and gradually enriched by adding \textit{Annotation} components. Besides spotting relevant excerpts (\textit{Annotate}), GPT can also help in summarizing the evaluation so far for the criterion at hand (\textit{Compile}) or asking for alternative viewpoints (\textit{Provide viewpoints})\footnote{The \textit{Recap} option is not prompt-based but just a rendering of the assessment of the criterion so far with the annotations and saved content (see Fig \ref{fig:viewpoint}.B)}. Fig. \ref{fig:viewpoint}  shows the case for \textit{Provide viewpoints}.


\subsection{Full Review}
A \textit{Review} is composed of multiple \textit{CriterionReview} components. \textit{AnnotateGPT} enables the creation of an editable review report using the \textit{Compilations} from the \textit{CriterionReview}, which are in turn derived from the \textit{Excerpts} and \textit{Comments} in the assembled \textit{Annotations}. This final review can be structured along with either the review criterion  or the sentiment. Functionalities \textit{reviewByCriteria} and \textit{reviewBySentiment} help to deliver these reports. As an example, in this URL\footnote{\href{https://rebrand.ly/annotateGPT\_Report}{ https://rebrand.ly/annotateGPT\_Report}} you can download the review report generated for this manuscript.

\section{Evaluation}\label{sec:tam}
This evaluation does not test the accuracy of GPT-4 in highlighting the right paragraphs. Assessing the quality of the annotations produced by GPT-4 is outside the scope of this work which takes as a premise the increasing accuracy that LLMs exhibit in this task. Rather, this work's main contribution should be sought in the interaction model for AI-human collaboration, and its realization in AnnotateGPT. It is then a question of technology acceptance, i.e., the extent reviewers perceive this model as ease of use and useful. We then resort to  the Technology Acceptance Model (TAM) \cite{davis1985}. 

\textbf{Goal.} \textbf{Assessing} the ease of use and usefulness of AnnotateGPT \textbf{with respect to} peer review \textbf{from the point of view of} human reviewers  \textbf{in the context of} criteria-based academic manuscript reviewing. 

\textbf{Execution.} Nine participants with experience in manuscript reviewing were locally selected.
Participants were initially asked to use \textit{AnnotateGPT} to review one of their own papers. This ensures participants were familiar with the content, enhancing their ability to understand \textit{AnnotateGPT}'s output. 

\begin{figure*}[t!]
    \centering
	\includegraphics[width=\linewidth]{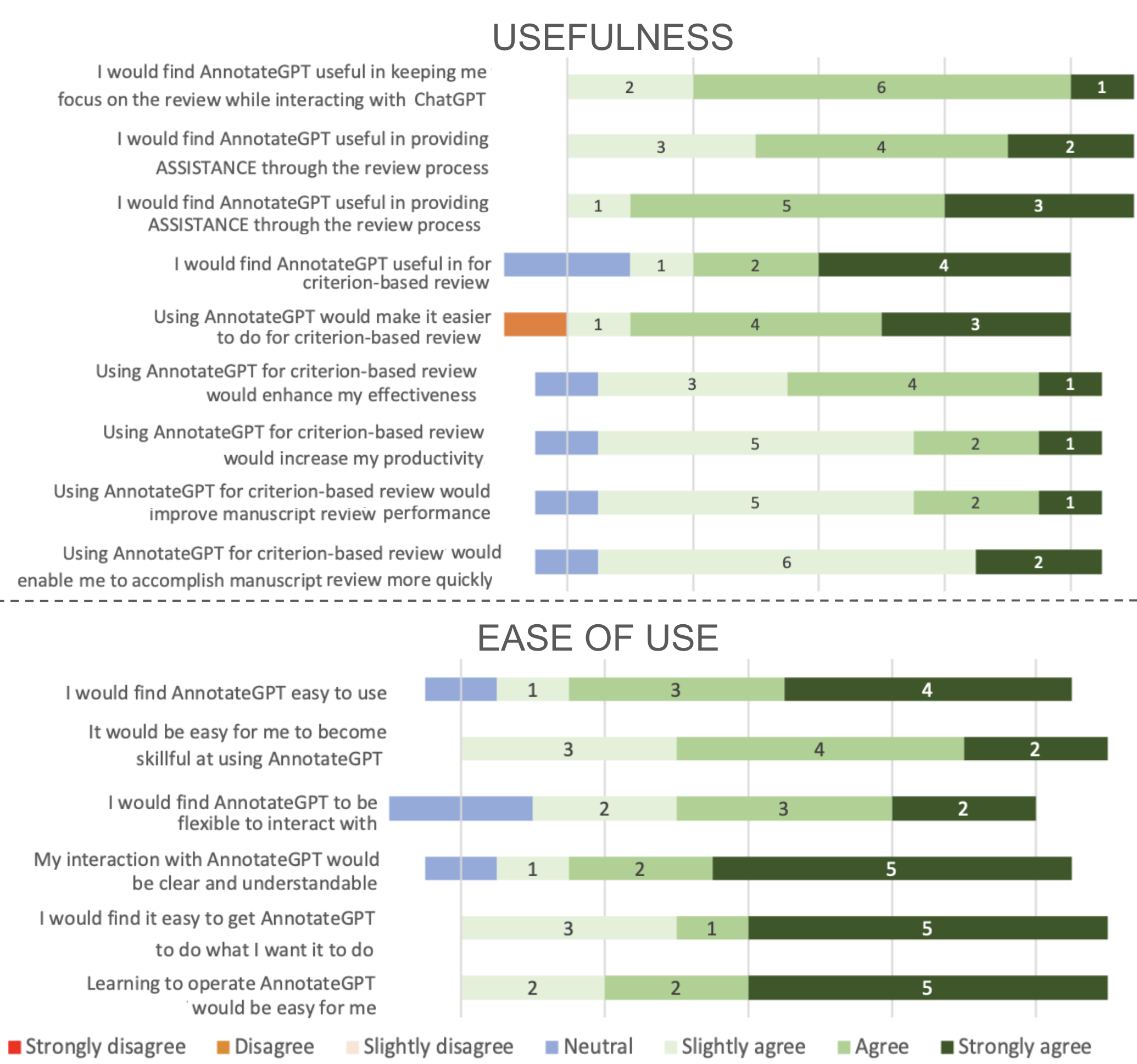}
	\caption{Perceived Usefulness \& Perceived Ease of use}
	\label{fig:tam}
\end{figure*}
\textbf{Results.} Fig. \ref{fig:tam} reveals a prevailing agreement among participants regarding the usefulness of \textit{AnnotateGPT}. However, there is a tendency to value more the role of \textit{AnnotateGPT} as a focus and criterion-consistency enabler rather than as a tool to increase productivity itself. As for `perceived ease of use' , results tend to be more conclusive on the seamlessness with which \textit{AnnotateGPT} is embedded within the PDF viewer. The results suggest that using \textit{AnnotateGPT} does not interfere with the existing GUI gestures of the viewer, hence reducing onboarding friction.

\subsection{Threats to validity}

The results demonstrated strong \textbf{construct validity} with Cronbach's   $\alpha$ values  of 0.8 and 0.83 for usefulness and ease of use, respectively. It is important to note that the assessment of 'usefulness' may be influenced by the GPT's ability to accurately identify the relevant excerpts. Although we made it clear beforehand, some participants might perceive the effectiveness of GPT as the determining factor, rather than solely considering the tool's usefulness in interacting with the LLM, regardless of the accuracy of the annotations. Regarding \textbf{internal validity}, we consider the subject's reviewing expertise and the manuscript's research topic to be confounding variables. Further evaluation is needed to determine the weight of these factors. Finally, in terms of \textbf{external validity}, manuscript annotation in domains other than IS can vary, reflecting the unique norms, values, and methodological approaches of each field. For example, in the humanities, peer review may focus more on argumentation, originality of perspective, and depth of analysis, while in the sciences, experimental design, statistical rigor, and reproducibility may be emphasized. However, these differences pertain to review criteria rather than the practice of annotation. As long as review criteria exist, the prompt identification of relevant excerpts as evidence of meeting these criteria is beneficial, regardless of the discipline.


\section{Formalization of Learning}\label{sec:generalization}

This section elaborates on the extent the insights of this project can be generalized. It specifically examines two facets: firstly, the generalization of the solution instance, i.e.,  the limitations of \textit{AnnotateGPT} as a \textit{solution} to performant review. Secondly, it explores the generalization of the problem instance, questioning to what extent \textit{performant feedback} is a \textit{problem} for  stakeholders other  than \textit{reviewers}.

\subsection{Generalization of the Solution Instance} 

\textit{AnnotateGPT} employs LLMs akin to content-based recommender systems. These recommendations, however, are subject to potential inaccuracies, including false positives (i.e., highlighting an excerpt not related to the criterion at hand) and false negatives (i.e., failing to spot an excerpt that relates to the criterion at hand). In the case of manuscript reviewing, the impact of false negatives is limited to cause some unnecessary fatigue in the reviewer's attention by highlighting unnecessary paragraphs. However, false negatives have a more significant impact as they could lead to overlooking relevant paragraphs. Therefore, \textit{AnnotateGPT} is limited by the precision of current LLMs. In order to mimic the advancements in recommender systems, \textit{AnnotateGPT} could provide assistance in various ways, specifically:
 \begin{itemize}
     \item leveraging user feedback. This can be achieved by allowing reviewers to provide feedback on the recommended paragraphs' relevance. By collecting this feedback, the system can iteratively refine its recommendations and adapt to each reviewer's preferences. This interactive feedback loop can lead to a personalized and fine-tuned paragraph recommendation system, further improving precision,
     \item improving the interpretability, i.e.,  generating explanations for the model's recommendations. These explanations can provide insights into why specific paragraphs were selected, helping reviewers understand and trust the system's recommendations. This aligns with the recent initiative for self-explained LLM models \cite{huang2023can}.
 \end{itemize}
The aforementioned limitations pertain to the LLM side. From a front-end perspective, \textit{AnnotateGPT} also has certain limitations due to its interaction metaphor. When reviewers engage with annotated manuscripts, they may become overly reliant on the highlighted excerpts and fail to engage with the full text, potentially missing important context or content that was not annotated. To address this issue, \textit{AnnotateGPT} offers the ability to toggle annotations on and off. 

 \subsection{Generalization of the Problem Instance} 
 \textit{AnnotateGPT} focuses on the reviewer as the stakeholder. However, manuscript feedback might also be `a problem' for both conference endowments and manuscript authors. 

\textbf{Conference organizers can benefit from using LLM to normalize and maintain the quality standards of their conferences.}  \textit{AnnotateGPT}  offers a variety of pre-set prompts for evaluating review criteria. These prompts can be customized to align with the specific requirements of the conference profile. These customized review platforms can be readily installed as browser plugins adjusting the current tool and distributed across PCs. This might help a more consistent review, ultimately enhancing the overall quality and prestige of the conference.

Furthermore, conference endowments are in a favorable position to utilize the abundance of reviews from past editions. As long as confidentiality issues are addressed with reviewers and authors, these reviews can be used for fine-tuning LLMs\footnote{LLM fine-tuning is a supervised learning process where you use a dataset of labeled examples to update the weights of LLM and make the model improve its ability for specific tasks}. This involves transforming general-purpose LLMs like GPT-4 into specialized tools that are specifically tuned for annotation-based reviews.

\textbf{Manuscript authors can benefit from a preliminary assessment.} Systems like ChatGPT might be used to generate the title, the abstract, a structured introduction that sets the context, or the keywords of the manuscript \cite{biswas2023chatgpt}. However, our content here is not about writing but about giving feedback. Regardless of how the manuscript is written (with the assistance of ChatGPT or not), LLMs can also be used to assist in the delicate task of giving feedback. The \textit{AnnotateGPT}'s contribution is not in generating content (``provide me with an introduction'') but in assessing content (``does \textit{this} introduction capture this manuscript's essence?''). Here, the LLM does not provide content but assess human-provided content. By highlighting what the LLM considers relevant excerpts, authors are encouraged to \textit{self-reflect} on their narratives and identify both weaknesses (to be addressed or explicitly acknowledged in the manuscript) and strengths (to be further developed in the manuscript). If authors use spell-check on their manuscripts before submitting them for review, why not use tools like \textit{AnnotateGPT} to help authors check in advance how well their drafts meet the review criteria set by the publication venue? So we did  for this very manuscript\footnote{You can download the report from the following URL, then, open the HTML file in a web browser: \href{https://rebrand.ly/annotateGPT\_Report}{ https://rebrand.ly/annotateGPT\_Report}}. It is not difficult to imagine that in the near future, authors will conduct a preliminary review to determine the extent to which they meet the venue's criteria. Reviewers are likely to appreciate this effort.

\section{Conclusions}
To keep up with authors empowered by AI, reviewers should also have access to similar tools. We have researched the use of AI as a supportive tool and propose a nuanced approach where AI aids reviewers in identifying important sections of the manuscript. To accomplish this, we recommend using "annotation" as a means to enhance collaboration between the AI agent and the reviewer. This vision is realized in \textit{AnnotateGPT}, a web-based PDF visualizer that combines criterion-driven prompts with language models like GPT-4.


We have different follow-ons in mind. In terms of rigor, we need larger quantitative evaluations (involving more subjects) and qualitative evaluations (involving conference endowment). In terms of relevance, our goal is to assess the effectiveness of open-source LLMs in pursuing reducing expenses (although AnnotateGPT is free, GPT-4 is not), while also exploring the fine-tuning of these models to enhance their effectiveness in  review. 

%
%
%
\bibliographystyle{splncs04}
\bibliography{sample}

\begin{thebibliography}{10}
\providecommand{\url}[1]{\texttt{#1}}
\providecommand{\urlprefix}{URL }
\providecommand{\doi}[1]{https://doi.org/#1}

\bibitem{HowtoMas25:online}
How to master reverse prompt engineering with chatgpt. \url{https://www.allabtai.com/how-to-master-reverse-prompt-engineering-with-chatgpt/}, (Accessed on 11/20/2023)

\bibitem{PeerRevi42}
AJE: Peer review: How we found 15 million hours of lost time, \url{https://www.aje.com/arc/peer-review-process-15-million-hours-lost-time/}, (Accessed on 11/24/2023)

\bibitem{biswas2023focus}
Biswas, S., Dobaria, D., Cohen, H.L.: {Focus: Big Data: ChatGPT and the Future of Journal Reviews: A Feasibility Study}. The Yale Journal of Biology and Medicine  \textbf{96}(3), ~415 (2023)

\bibitem{biswas2023chatgpt}
Biswas, S.S.: {ChatGPT for Research and Publication: A Step-by-Step Guide}. The Journal of Pediatric Pharmacology and Therapeutics  \textbf{28}(6),  576--584 (2023)

\bibitem{Boud2013}
Boud, D., Molloy, E.: Rethinking models of feedback for learning: the challenge of design. Assessment \& Evaluation in higher education  \textbf{38}(6),  698--712 (2013). \doi{10.1080/02602938.2012.691462}

\bibitem{checco2021ai}
Checco, A., Bracciale, L., Loreti, P., Pinfield, S., Bianchi, G.: {AI}-assisted peer review. Humanities and Social Sciences Communications  \textbf{8}(1),  1--11 (2021)

\bibitem{davis1985}
Davis, F.: {A Technology Acceptance Model for Empirically Testing New End-User Information Systems}  (1985)

\bibitem{enholm2022artificial}
Enholm, I.M., Papagiannidis, E., Mikalef, P., Krogstie, J.: Artificial intelligence and business value: A literature review. Information Systems Frontiers  \textbf{24}(5),  1709--1734 (2022)

\bibitem{ghosal2019deepsentipeer}
Ghosal, T., Verma, R., Ekbal, A., Bhattacharyya, P.: {DeepSentiPeer: Harnessing sentiment in review texts to recommend peer review decisions}. In: Proceedings of the 57th Annual Meeting of the Association for Computational Linguistics. pp. 1120--1130 (2019)

\bibitem{huang2023can}
Huang, S., Mamidanna, S., Jangam, S., Zhou, Y., Gilpin, L.H.: Can large language models explain themselves? a study of llm-generated self-explanations. arXiv preprint arXiv:2310.11207  (2023)

\bibitem{lin2023automated}
Lin, J., Song, J., Zhou, Z., Chen, Y., Shi, X.: Automated scholarly paper review: Concepts, technologies, and challenges. Information Fusion p. 101830 (2023)

\bibitem{Nicol2010}
Nicol, D.: {From monologue to dialogue: improving written feedback processes in mass higher education}. Assessment {\&} Evaluation in Higher Education  \textbf{35}(5),  501--517 (aug 2010). \doi{10.1080/02602931003786559}

\bibitem{OpenAI2023ChatGPT}
OpenAI: {ChatGPT (Nov 30 version}. Large language model (2023), \url{https://chat.openai.com/chat}, accessed: 2023-11-08

\bibitem{park2023use}
Park, S.H.: {Use of generative artificial intelligence, including large language models such as ChatGPT, in scientific publications: policies of KJR and prominent authorities}. Korean Journal of Radiology  \textbf{24}(8), ~715 (2023)

\bibitem{donnel2004}
Porter-O'Donnell, C.: Beyond the yellow highlighter: Teaching annotation skills to improve reading comprehension. The English Journal  \textbf{93}(5),  82--89 (2004). \doi{10.2307/4128941}

\bibitem{PublonsG57}
Publons: Global state of peer review. \url{https://publons.com/static/Publons-Global-State-Of-Peer-Review-2018.pdf}, (Accessed on 11/24/2023)

\bibitem{spyns2015scientific}
Spyns, P., Vidal, M.E.: Scientific Peer Reviewing: Practical Hints and Best Practices. Springer (2015)

\bibitem{srivastava2023day}
Srivastava, M.: {A day in the life of ChatGPT as an academic reviewer: Investigating the potential of large language model for scientific literature review}  (2023)

\bibitem{tennant2017multi}
Tennant, J.P., Dugan, J.M., Graziotin, D., et~al.: A multi-disciplinary perspective on emergent and future innovations in peer review. F1000Research  \textbf{6} (2017)

\bibitem{wang2020reviewrobot}
Wang, Q., Zeng, Q., Huang, L., Knight, K., Ji, H., Rajani, N.F.: Reviewrobot: Explainable paper review generation based on knowledge synthesis. arXiv preprint arXiv:2010.06119  (2020)

\bibitem{Ware2011}
Ware, M.: Peer review: Recent experience and future directions. New Review of Information Networking  \textbf{16}(1),  23--53 (2011). \doi{10.1080/13614576.2011.566812}, \url{https://doi.org/10.1080/13614576.2011.566812}

\bibitem{ware2011peer}
Ware, M.: Peer review: Recent experience and future directions. New Review of Information Networking  \textbf{16}(1),  23--53 (2011)

\bibitem{yuan2022can}
Yuan, W., Liu, P., Neubig, G.: Can we automate scientific reviewing? Journal of Artificial Intelligence Research  \textbf{75},  171--212 (2022)

\end{thebibliography}

\end{document}